\documentclass[final,5p]{elsarticle}

\usepackage{amssymb,amsmath}
\usepackage{url}
\usepackage{bbm}
\usepackage{color}
\usepackage{wrapfig,lipsum,booktabs}
\usepackage{graphicx}
\usepackage{float}
\usepackage{multirow}
\usepackage{makecell}
\usepackage{array}
\usepackage{diagbox}
\usepackage{arydshln}
\usepackage{stfloats}
\usepackage{listings}
\usepackage{pifont}
\usepackage[dvipsnames]{xcolor}

\usepackage{algorithm}
\usepackage{algorithmicx}
\usepackage[noend]{algpseudocode}

\newif\ifshowrevisions
\showrevisionsfalse 

\newtheorem{definition}{Definition}

\newcommand{\com}{\textsf{com}}
\newcommand{\prf}{\textsf{prf}}
\newcommand{\grd}{\textsf{grd}}
\newcommand{\stb}{\textsf{stb}}

\journal{Neurocomputing}

\begin{document}

\begin{frontmatter}

\title{Enhancing Conflict Resolution in Language Models via Abstract Argumentation}

\author[inst1,inst3]{Zhaoqun Li}
\author[inst2]{Xiaotong Fang}
\author[inst1,inst3]{Chen Chen}
\author[inst2]{Mengze Li}
\author[inst1,inst3]{Beishui Liao}

\affiliation[inst1]{organization={School of Philosophy, Zhejiang University},
                    country={China}}
\affiliation[inst2]{organization={College of Computer Science and Technology, Zhejiang University},
                    country={China}}
\affiliation[inst3]{organization={The State Key Lab of Brain-Machine Intelligence},
                    country={China}}

\begin{abstract}
In recent years, large language models (LLMs) have made significant advancements in developing human-like and engaging dialogue systems. 
However, in tasks such as consensus-building and persuasion, LLMs often struggle to resolve conflicts arising from incomplete or inconsistent information, revealing their limitations in real-world applications. 
Given these limitations, abstract argumentation, a specialized logical framework designed to resolve conflicts and inconsistencies, becomes particularly relevant. 
In this paper, we aim to enhance the conflict-solving capabilities of LLMs by leveraging formal abstract argumentation, integrating language model learning with symbolic computation.
To achieve this, we develop and curate a dataset comprising diverse abstract argumentation frameworks, accompanied by detailed explanations of the argument acceptability computation process. 
Subsequently, we fine-tune LLMs on this dataset, focusing on abstract conflict resolution tasks. 
As a comparative baseline, LLMs are also evaluated using a chain-of-thought approach, however, they fail to solve the conflict-based arguments effectively.
Our experiments demonstrate that process explanations play a crucial role in learning. Models trained with explanations exhibit superior generalization accuracy compared to those trained solely on question-answer pairs. 
Furthermore, leveraging LLMs' self-explanation capabilities, our approach provides detailed illustrations that mitigate the lack of transparency typically associated with neural networks. 
\end{abstract}

\end{frontmatter}

\section{Introduction}

Recent advancements in large language models (LLMs) have significantly improved their proficiency in addressing a range of complex reasoning problems, thereby demonstrating their adaptability across diverse domains~\cite{fang2024large,huang2022language,alon2022neuro,pan-etal-2023-logic}. 
Nevertheless, LLMs exhibit comparatively limited efficacy in formal logical reasoning tasks~\cite{han2022folio,li2025exploring,eisape2024systematic,cheng2025empowering}, frequently resulting in hallucinations and derivation errors.
These characteristics render them unreliable in applications such as conflict resolution and mediation.
Therefore, it is desirable to enhance LLMs' ability to perform reasoning and make relatively sound decisions when dealing with complex and conflicting information.

Argumentation is a formal form of defeasible reasoning that plays a crucial role in various real-world applications, such as legal decision-making~\cite{prakken2023formal,chi2021optimized}, policy compliance analysis~\cite{dong2022defeasible}, and ethical decision-making~\cite{awad2020approach}. 
Combining argumentation with deep learning and gaining insights into how humans engage in argumentative reasoning is a vital aspect of artificial intelligence development.
Previous research has pioneered this area by employing neural networks to compute acceptability semantics within the abstract argumentation framework (AAF)~\cite{dung1995acceptability,malmqvist2024approximating,Thorburn2022OptimizingLM}. 
Notable contributions include AGNN and EGNN ~\cite{ijcai2020p231,craandijk2022enforcement}, which explores the use of graph neural network (GNN) to determine argumentation semantics, demonstrating that neural networks can effectively learn and predict the acceptability of arguments in various scenarios. 
These works highlight the promising intersection of neural networks and argumentation theory, however, the intrinsic black-box characteristic of neural networks leads to the lack of explainability in prediction.
It is hard to show explicitly how their models predict the acceptability as their basis for decision-making, which is crucial in real applications.

AAF justifies argument acceptability through different semantics, thereby making selections among conflicting sets of arguments. 
Consequently, semantics computation is of great importance.
In this paper, we focus on enhancing the conflict-solving capabilities of LLMs by leveraging abstract argumentation, integrating language model training with symbolic computation.
Our objectives are \textit{twofold}: first, to train an LLM capable of recognizing argumentation frameworks and assessing argument acceptability; second, to enhance the explainability of LLMs through this training, thereby potentially addressing some of their inherent limitations.
To achieve these goals, we leverage the self-explanation capacity of LLMs~\cite{huang2023large,chuang2024faithlm} and train language models on a curated data corpus using supervised fine-tuning (SFT).
Furthermore, we incorporate reinforcement learning from human feedback (RLHF) to further improve the interpretability and pedagogical quality of the model’s explanations~\cite{ouyang2022training,bakker2022fine,wong2024aligning}.
This approach enables LLM to interpret its behavior during the problem-solving process through text generation. 
Specifically, we establish an AAF benchmark comprising formal frameworks that represent various conflict types.
On the benchmark, we explore two fundamental tasks: computing the grounded extensions and complete extensions based on labelling algorithms.
In the dataset, each sample comprises an AAF described using a graph description language, with the execution information of a recursive algorithm and justification process attached as the explanation. 
We then fine-tune two foundational models, Llama-3~\cite{llama3} and Qwen2.5~\cite{qwen25}, on our established dataset by SFT and RLHF.
The data organization embeds a post-explainability mechanism into the training process, which can reflect the LLM's capability in understanding symbolic computation.

In the experiment, we demonstrate that our approach marginally improves accuracy compared to zero-shot prompting without fine-tuning. 
Through concrete error analysis, we enumerate common inference mistakes in the semantics computation and identify directions for further improvement.
Additionally, we show that process explanation is crucial for argumentation computation in LLMs. 
Models trained with process explanations exhibit superior generalization accuracy compared to those trained solely on question-answer data. 
By leveraging the self-explanation capabilities of LLMs, our approach provides detailed illustrations that mitigate the lack of transparency typically associated with neural network mechanisms. 

To summarize, our contributions are as follows:
\begin{itemize}
    \item We establish an argumentation benchmark with algorithm explanation, enabling the training of explainable LLMs to argument acceptability using labeling algorithms.
    \item A methodology for integrating argumentation frameworks with language models is proposed through fine-tuning, improving the model's ability to solve conflict and explain underlying computation dynamics.
    \item We show that training LLMs with process explanations has higher accuracy and transparency over zero-shot prompting, facilitating applications in real-world decision-making scenarios.
\end{itemize}

\section{Related Work}

Within the domain of formal logical deduction using language models, the primary focus has predominantly been on first-order logic~\cite{tian2021diagnosing}. 
Works such as RuleTaker~\cite{Clark2020Transformers} and FOLIO~\cite{han2022folio} exemplify the application of artificial intelligence in deducing logical conclusions from structured premises, thereby demonstrating models' capacity for formal logical reasoning. 
However, argumentative reasoning, which involves both logical reasoning and graph comprehension, has been largely overlooked~\cite{guo2023gpt4graph}.

Previous research has attempted to integrate argumentation frameworks or with neural networks for interpretation~\cite{potyka2021interpreting} and semantics computation~\cite{craandijk2022enforcement,ijcai2020p231}. 
For instance, the work on interpreting neural networks as quantitative argumentation frameworks~\cite{potyka2021interpreting} explores how neural networks can be understood through the lens of argumentation. 
Similarly, the AGNN~\cite{ijcai2020p231} investigates the application of deep learning techniques to abstract argumentation semantics. 
Additionally, research in~\cite{craandijk2022enforcement} proposes enforcement heuristics for argumentation with deep reinforcement learning, aiming to enhance the decision-making process in argumentation frameworks.
In~\cite{mileva2023unifying}, researchers aim to consolidate diverse approaches to learning argumentation semantics within a unified framework.

Recent research has integrated argumentation techniques with large language models to enhance reasoning capabilities in various tasks.  
Görur et al.~\cite{gorur2024large} demonstrate that LLMs like Llama-2 and Mistral outperform traditional models in Relation-Based Argument Mining by effectively identifying support and attack relations between arguments.  
Chen et al.~\cite{chen2024exploring} conduct a comprehensive investigation into the potential of LLMs, such as ChatGPT, for computational argumentation. 
Their study focuses on argument mining and generation tasks, evaluating these models in both zero-shot and few-shot learning scenarios across diverse datasets.  
Freedman et al.~\cite{freedman2024argumentative} propose an innovative approach to augment the reasoning abilities of LLMs through the incorporation of argumentative reasoning, which leverages LLMs to construct argumentation frameworks.  
The integration of these methodologies holds significant promise for advancing the development of robust artificial intelligence systems capable of sophisticated argumentation and reasoning.

\section{Preliminaries}
\label{sec:pre}

\begin{table*}
\centering
\scalebox{1}{
\begin{tabular}{|c|cccccccc|cccc|} 
\hline
Labelling   &1     &2     &3     &4     &5     &6     &7     &8     
& \com   &\grd &\prf & \stb \\
\hline
$\mathcal{L}_1$   
&$\mathtt{IN}$  &$\mathtt{OUT}$ &$\mathtt{UNDEC}$   &$\mathtt{OUT}$ &$\mathtt{UNDEC}$   &$\mathtt{IN}$  &$\mathtt{IN}$  &$\mathtt{UNDEC}$
&\checkmark &\checkmark &\ding{55}  &\ding{55} \\
$\mathcal{L}_2$ 
&$\mathtt{IN}$  &$\mathtt{OUT}$ &$\mathtt{IN}$  &$\mathtt{OUT}$ &$\mathtt{OUT}$ &$\mathtt{IN}$  &$\mathtt{IN}$  &$\mathtt{OUT}$
&\checkmark &\ding{55}  &\checkmark &\checkmark \\
$\mathcal{L}_3$
&$\mathtt{IN}$  &$\mathtt{OUT}$ &$\mathtt{OUT}$ &$\mathtt{OUT}$ &$\mathtt{IN}$  &$\mathtt{IN}$  &$\mathtt{IN}$  &$\mathtt{IN}$
&\checkmark &\ding{55}  &\checkmark &\checkmark \\
\hline
\end{tabular}
}
\caption{Possible complete labellings of the argumentation framework from Figure~\ref{fig_aaf1}.}
\label{tab_label} 
\end{table*}

In the context of AAF, arguments are considered atomic entities devoid of internal structure and only a binary attack relation is considered, allowing for a highly adaptable approach to argumentation. 
The acceptability semantics of each model within the framework is determined based on the interactions among arguments.

\begin{definition}
An abstract argumentation framework $\mathcal{F}$ is a tuple $\mathcal{F}=(\mathcal{A}, \mathcal{R})$ where $\mathcal{A}$ is a set of arguments and $\mathcal{R}$ is a relation $\mathcal{R} \subseteq \mathcal{A}\times \mathcal{A}$.
\end{definition}

For two arguments $a, b \in \mathcal{A}$, the relation $(a, b) \in \mathcal{R}$ signifies that argument $ a $ attacks argument $ b $. For a set $ S \subset \mathcal{A} $ and an argument $ a $, we say that $ S $ attacks $ a $ if there exists $ b \in S $ such that $(b, a) \in \mathcal{R}$; 
similarly, $ a $ is said to attack $ S $ if there exists $ b \in S $ such that $(a, b) \in \mathcal{R}$.  
Argument $a$ is defended by set S if and only if for every $b \in \mathcal{A}$ such that $(b,a)\in\mathcal{R}$ there exists $c \in S$ such that $c$ attacks $b$.
We further state that $ S $ attacks a set $ P $ if there exist $ a \in S $ and $ b \in P $ such that $ a $ attacks $ b $.

\begin{figure}[t]
    \centering
    \includegraphics[width=0.6\linewidth]{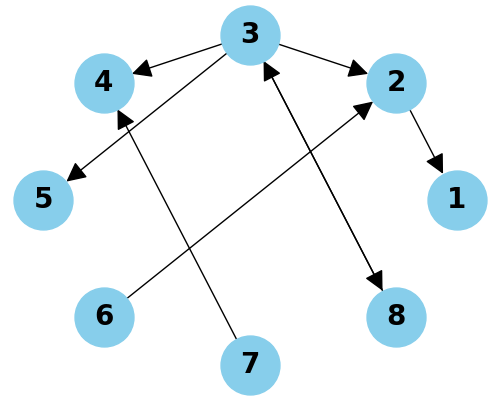}
    \caption{Example abstract argumentation framework.}
    \label{fig_aaf1}
\end{figure}

Abstract argumentation frameworks can be concisely represented by directed graphs, where arguments are nodes and edges represent the attack relation.
We use the AAF in Figure~\ref{fig_aaf1} (denoted as Example~\ref{fig_aaf1}) as our illustration example in this paper. 
The status of a given argument is determined through argumentation semantics, often yielding results in the form of extensions~\cite{dung1995acceptability} or labellings~\cite{wu2010labelling}, which can be used interchangeably.
In this work, we primarily use labellings, and the explanation of computation also involves concepts such as conflict-freeness.

\begin{figure*}[t]
    \centering
    \includegraphics[width=1\linewidth]{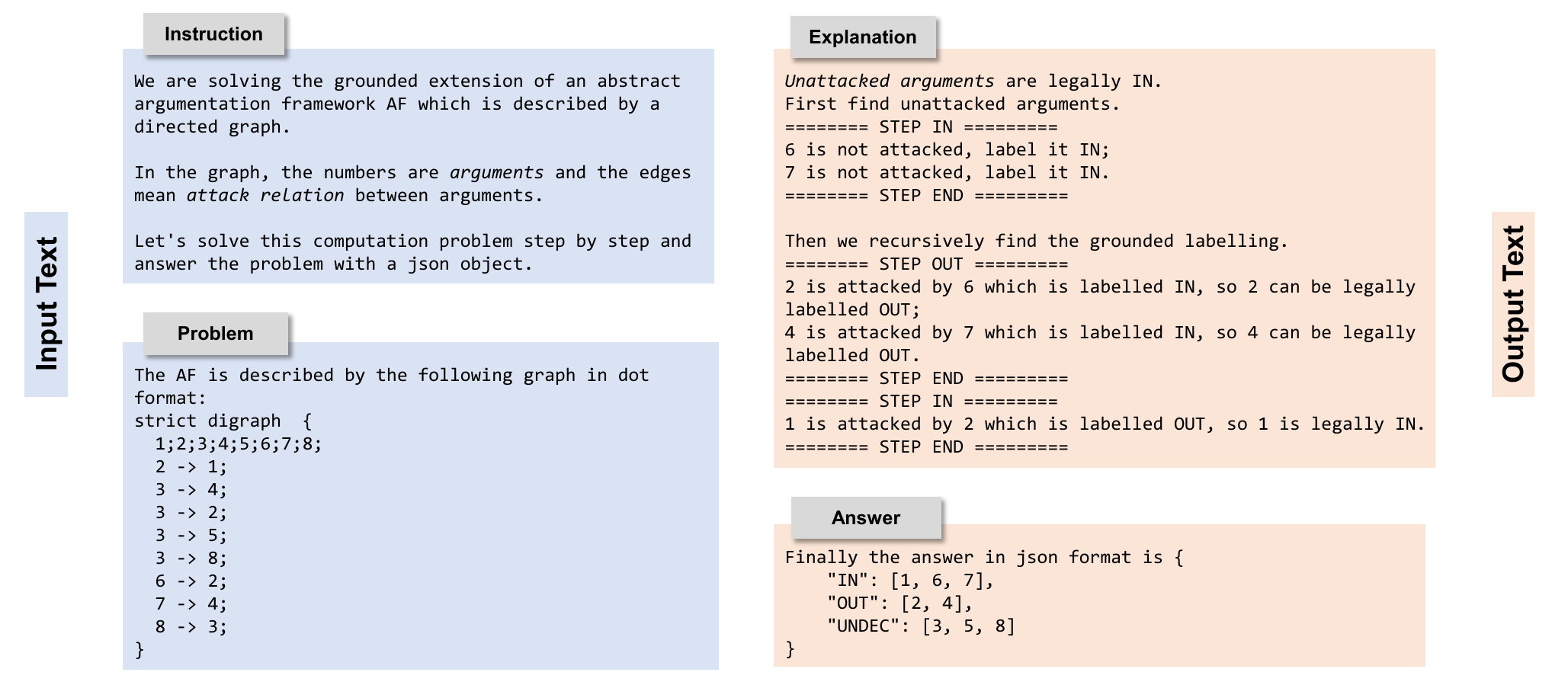}
    \caption{Data formalization of solving the grounded labelling of Example~\ref{fig_aaf1}. The explanation involves the computation process through recursive IN and OUT steps. In the text, we use italics to emphasize important concepts in AAF.}
    \label{fig_data1}
\end{figure*}

\begin{definition}
A labelling $\mathcal{L}$ for an abstract argumentation framework $\mathcal{F} =(\mathcal{A}, \mathcal{R})$ is a function $\mathcal{L} : \mathcal{A} \rightarrow$ \{$\mathtt{IN}$, $\mathtt{OUT}$, $\mathtt{UNDEC}$\}.
\end{definition}

A labelling $\mathcal{L}$ assigns to each argument $a \in \mathcal{A}$ one of the following values: $\mathtt{IN}$, $\mathtt{OUT}$, or $\mathtt{UNDEC}$, indicating that the argument is accepted, rejected, or undecided, respectively. 
Let $\mathtt{IN}(\mathcal{L}) = \{a \mid \mathcal{L}(a) = \mathtt{IN}\}$ and define $\mathtt{OUT}(\mathcal{L})$ and $\mathtt{UNDEC}(\mathcal{L})$ analogously. 
The set $\mathtt{IN}(\mathcal{L})$ for a labelling $\mathcal{L}$ under a semantics $\sigma$ is a $\sigma$-extension.

\begin{definition}
\label{def_legal}
Let $\mathcal{L}$ be a labelling for $\mathcal{F}=(\mathcal{A}, \mathcal{R})$ and $a,b\in\mathcal{R}$. 
\begin{itemize}
\item $a$ is legally $\mathtt{IN}$ iff $a$ is labelled $\mathtt{IN}$ and every $b$ that attacks $a$ is labelled $\mathtt{OUT}$.
\item $a$ is legally $\mathtt{OUT}$ iff $a$ is labelled $\mathtt{OUT}$ and there is at least one $b$ that attacks $a$ and $b$ is labelled $\mathtt{IN}$.
\item $a$ is legally $\mathtt{UNDEC}$ iff $a$ is labelled $\mathtt{UNDEC}$ and not every $b$ that attacks $a$ is labelled $\mathtt{OUT}$, and there is no $b$ that attacks $a$ such that $b$ is labelled $\mathtt{IN}$.
\end{itemize}
\end{definition}

There are various semantics. In this work, we consider the complete, grounded, preferred, and stable semantics, denoted by \com, \grd, \prf, and \stb, respectively.

\begin{definition}
Let $\mathcal{L}$ be a labelling, we call $\mathcal{L}$ 
\begin{itemize}
\item an admissible labelling iff in $\mathcal{L}$ no arguments are illegally $\mathtt{IN}$ and no arguments are illegally $\mathtt{OUT}$;
\item a complete labelling iff $\mathcal{L}$ is an admissible labelling without arguments that are illegally $\mathtt{UNDEC}$;
\item a grounded labelling iff there there does not exist a complete labelling $\mathcal{L}^\prime$ such that $\mathtt{IN}(\mathcal{L}^\prime)\subset\mathtt{IN}(\mathcal{L})$;
\item a preferred labelling iff there does not exist a complete labelling $\mathcal{L}^\prime$ such that $\mathtt{IN}(\mathcal{L}^\prime)\supset\mathtt{IN}(\mathcal{L})$;
\item a stable labelling iff $\mathtt{UNDEC}(\mathcal{L}) = \emptyset$.
\end{itemize}
\end{definition}

Grounded labelling always exists and is unique. 
In contrast, other semantics may exhibit dependency on the specific structure of the argumentation framework.
We show all possible complete labellings of Example~\ref{fig_aaf1} in Table~\ref{tab_label}.

\section{Dataset Construction}
\label{sec:data}

\begin{figure*}[t]
    \centering
    \includegraphics[width=1\linewidth]{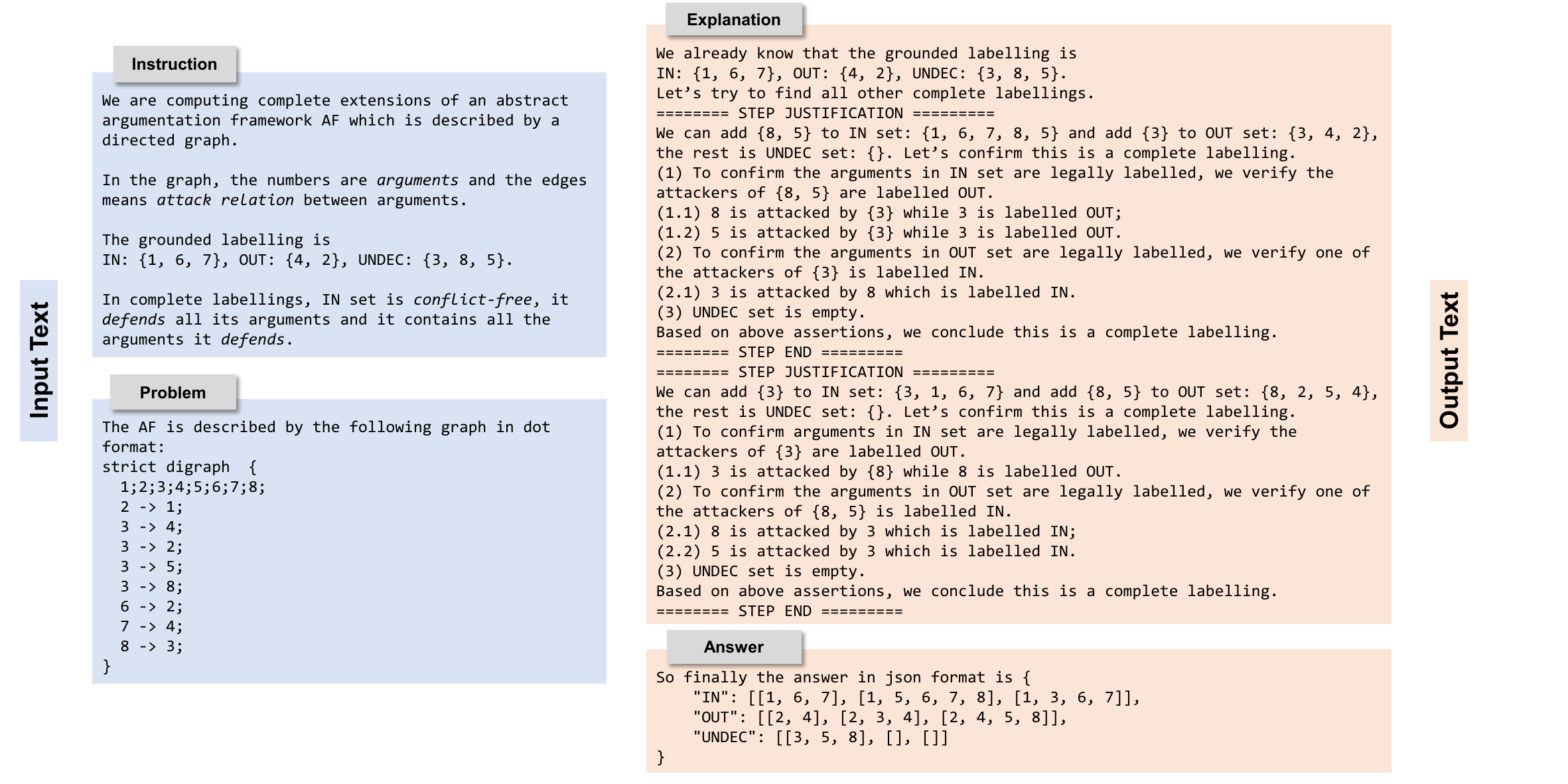}
    \caption{Data formalization of solving the complete labellings of Example~\ref{fig_aaf1}. 
    The explanation involves justification steps that verify each predicted complete labelling is legal.}
    \label{fig_data2}
\end{figure*}

This work focuses on leveraging semantics computation with LLMs for conflict resolution, rather than improving argumentation computation methods.
To construct a benchmark with algorithm explanations, we first generate randomly a variety of challenging argumentation frameworks following the methodology described in~\cite{ijcai2020p231}. 
In generation process, duplicate samples are removed and the ground truth semantics are computed. 
The number of arguments in an AAF ranges from 6 to 25. 
To encompass various levels of difficulty, we generate 3,000 samples for the training set and 100 samples for the test set for each number of arguments $n$, resulting in a total of 60,000 training samples and 2,000 test samples~\footnote{\url{https://github.com/dxbdxx/llm_af_computation}}.
The grounded, preferred, stable, and complete semantics have an average of 1.0, 1.4, 1.1, and 2.4 extensions per AAF, respectively, with 4.3, 6.4, 7.6, and 6.0 arguments per extension, respectively.

Designed for LLM supervised fine-tuning, each data sample follows an instruction-tuning data template that includes four components: instruction, problem, explanation, and answer. 
During training, the instruction and problem serve as input text, while the explanation and answer constitute the output text.
Figures~\ref{fig_data1} and~\ref{fig_data2} illustrate the specific data contents for solving the grounded extension and the complete extensions of Example~\ref{fig_aaf1}, respectively.
We define two \textit{basic tasks}: computing the grounded semantics and the complete semantics. 
It should be noted that the complete labellings are computed under the assumption that the grounded labelling is already known.
In practical applications, the computation involves a two-step process.

After predicting all complete labellings, 
preferred and stable labellings can be straightforwardly inferred. 
In preferred labellings, the $\mathtt{IN}(\mathcal{L})$ sets are maximal, and stable labellings are complete labellings in which no argument is $\mathtt{UNDEC}$.

\subsection{Instruction and Problem}
The instruction includes a basic introduction of AAF, the task description, and return format requirements. 
Since we are conducting formal reasoning, we ask LLMs to generate a JSON object at the end for clarity and concreteness. 
To render the abstract representation readable for both humans and LLMs, we adopt graph description languages to represent the original argumentation frameworks. 
We use three graph description languages: GraphML~\cite{brandes2002graphml}, Graphviz DOT, and JSON. 
For each data sample, one format is randomly chosen during the problem statement generation.

\subsection{Process Explanation}
For the computation of extensions, we detail the process of executing labelling algorithms to identify the correct labellings of AAF semantics. 
The explanation includes a step-by-step description of how we solve the problem, reflecting the principles of argumentative computation.

\subsubsection{Grounded Semantics}
The labelling algorithm for generating the grounded labelling begins by assigning $\mathtt{IN}$ to all arguments that are not attacked. 
It then proceeds iteratively: $\mathtt{OUT}$ is assigned to any argument attacked by an argument that has just been made $\mathtt{IN}$, and $\mathtt{IN}$ is assigned to arguments whose attackers are all $\mathtt{OUT}$. 
The iteration continues until no new arguments can be assigned $\mathtt{IN}$ or $\mathtt{OUT}$. Any arguments that remain unlabelled are then assigned $\mathtt{UNDEC}$. 
This algorithm is sound and complete as it effectively mimics the construction of the grounded extension through the iteration of a framework’s characteristic function~\cite{modgil2009proof}.

\subsubsection{Complete Semantics}
By definition, complete extensions are supersets of the grounded extension. 
Based on the grounded extension, an algorithm to compute all complete extensions involves changing the labels of some arguments in $\mathtt{UNDEC}(\mathcal{L_\grd})$ to $\mathtt{IN}$ or $\mathtt{OUT}$.
In this paper, we regard the computation of complete semantics as a two-step reasoning task that first computes the grounded semantics.
Then, instead of using a forward algorithm, we allow the LLM to directly predict each complete labelling $\mathcal{L}$. 
After that, we verify that the predicted labelling is a complete labelling by Definition~\ref{def_legal}. 

In the explanation, we provide all ground truth complete labellings and allow the LLM to implicitly learn how to compute them based on the grounded labelling. 
This annotation approach enables the LLM to learn to infer each complete extension and then verify it.

\subsection{Answer}
In the answer, we give the final output labellings in a JSON object, which contains the sets $\mathtt{IN}(\mathcal{L})$, $\mathtt{OUT}(\mathcal{L})$, and $\mathtt{UNDEC}(\mathcal{L})$.
The arguments in the sets are sorted by their number to avoid ambiguity.

\subsection{Enhancing Data Diversity}
To further increase data diversity and reduce overfitting, we utilize GPT-4o~\cite{gpt4} to paraphrase and polish the natural language descriptions in both training and testing samples. 
For each instance, the instruction and explanatory texts are automatically rephrased into multiple diverse versions, while the formal graph representation (such as DOT, GraphML, or JSON format) and the answer remain unchanged.

\section{Training and Baseline Methods}

In this section, we introduce the training strategies and baseline methods used for argumentation semantics computation with LLMs.

\subsection{Supervised Fine-Tuning}
\label{sec:sft}

We mainly utilize the constructed benchmark dataset to perform SFT of LLMs for semantics computation. 
Each data instance follows an instruction-tuning format, where the instruction and problem fields are used as input, and the explanation (optional) and answer fields as supervised output. 
Following standard instruction tuning practices, we adopt a sequence-to-sequence training paradigm, enabling the model to generate both step-by-step reasoning and the final solution in a structured JSON format.

\subsection{Chain-of-Thought Baseline}
Prompting generally results in models making predictions with lower accuracy compared to models that have been fine-tuned with substantial training data~\cite{zhang2023large}. 
To validate this observation and establish a baseline method, we design a Chain-of-Thought (CoT) pipeline~\cite{NEURIPS2022_9d560961} that computes extensions based on a labelling algorithm. 
The argumentation computation is divided into multiple steps, framed as a reasoning task.

For computing grounded semantics, LLMs are taught to identify arguments and their relations, starting with those that aren't attacked. They follow steps to label arguments until no more changes are needed.
As for complete Semantics, the models take arguments that are labelled $\mathtt{UNDEC}$ and decide which can be further add to set $\mathtt{IN}$ to form a complete extension, checking if their choices make sense.
The concrete prompts and implementation steps can be found in Appendix.

\subsection{Integrating Human Feedback}

The semantics computation based on labelling algorithms is primarily designed from a programmatic perspective.
In contrast, when humans manually compute extensions, they often adopt a more intuitive and natural approach to explanation. 
Human explanations tend to be expressed in a way that is easier to understand.
However, models trained solely via SFT typically generate outputs that closely follow the templates used during training, resulting in less flexible explanations.

To investigate the impact of human annotation on explanation quality, we design an experiment that incorporates reinforcement learning.
Specifically, we provide both the problem statements and reference answers to expert annotators, asking them to play the role of a teacher and annotate step-by-step explanations as if instructing a student on how to compute extensions.
The goal is to collect high-quality, pedagogical explanations that could potentially improve the explainability of model outputs.
In this paper, we annotate grounded semantics and a portion of the data.
More implementation details and annotated examples can be found in the Appendix.
\section{Experiment}
\label{sec:exp}


\subsection{Implementation Details}

In our training process, we employ the LoRA technique~\cite{hu2022lora} to facilitate efficient fine-tuning.
Throughout all experiments, we utilize the bfloat16 (bf16) computation datatype.
For model training, the maximum context length is set to 2048 tokens.
The learning rate is set to $1\text{e}{-}4$ for SFT, and the training time is around 10 hours.
For RLHF, the learning rate is $1\text{e}{-}6$.
The LLMs are trained on a server with eight NVIDIA A100-SXM-40GB GPUs.
For inference, we use vLLM~\cite{kwon2023efficient} and set the temperature to 0.3.
It takes 0.4 seconds to generate all responses for semantics computation on a single GPU.

\begin{table*}[t]
\begin{center}
\scalebox{0.93}[0.95]{
\begin{tabular}{lcccccccccccccc}
\toprule
\multirow{2}{*}{Models} & \multicolumn{2}{c}{\grd} && \multicolumn{3}{c}{\com} && \multicolumn{3}{c}{\prf} &&\multicolumn{3}{c}{\stb} \\
\cline{2-3} \cline{5-7} \cline{9-11} \cline{13-15} 
&ACC$_\sigma$   &MCC$_{s,c}$   &&ACC$_\sigma$   &MCC$_{s}$   &MCC$_{c}$   &&ACC$_\sigma$   &MCC$_{s}$   &MCC$_{c}$   &&ACC$_\sigma$   &MCC$_{s}$   &MCC$_{c}$
\cr
\hline
Llama3-8B & 0.18 & 0.17 && 0.08  & 0.41 & 0.31 && 0.10  & 0.31 & 0.31 && 0.24  & 0.21 & 0.28 \\
Llama3-70B & 0.37 & 0.42 && 0.28  & 0.89 & 0.66 && 0.39  & 0.85 & 0.66 && 0.39  & 0.43 & 0.54 \\
Qwen2.5-7B & 0.31 & 0.36 && 0.43  & 0.84 & 0.61 && 0.37  & 0.66 & 0.61 && 0.53  & 0.47 & 0.65 \\
Qwen2.5-72B  & 0.48 & 0.69 && 0.51  & 0.96 & 0.78 && 0.54  & 0.85 & 0.78 && 0.65  & 0.57 & 0.76 \\
GPT-3.5  & 0.15 & 0.37 && 0.17  & 0.85 & 0.52 && 0.29  & 0.67 & 0.52 && 0.30  & 0.35 & 0.34 \\ 
GPT-4o    & 0.54 & 0.73 && 0.48  & 0.98 & 0.64 && 0.49  & 0.74 & 0.64 && 0.58  & 0.52 & 0.58 \\
\hline
\specialrule{0em}{1pt}{1pt}
Qwen2.5-SFT & 0.97 & 0.99 && 0.91  & 0.99 & 0.96 && 0.88  & 0.98 & 0.96 && 0.90  & 0.97 & 0.95 \\
Llama3-SFT & 0.95 & 0.99 && 0.85  & 0.99 & 0.89 && 0.86  & 0.90 & 0.89 && 0.88  & 0.88 & 0.89 \\
\bottomrule
\end{tabular}}
\end{center}
\caption{The performance comparison of LLMs under different semantics.}
\label{tab_main}
\end{table*}

\begin{figure}[t]
    \centering
    \includegraphics[width=1\linewidth]{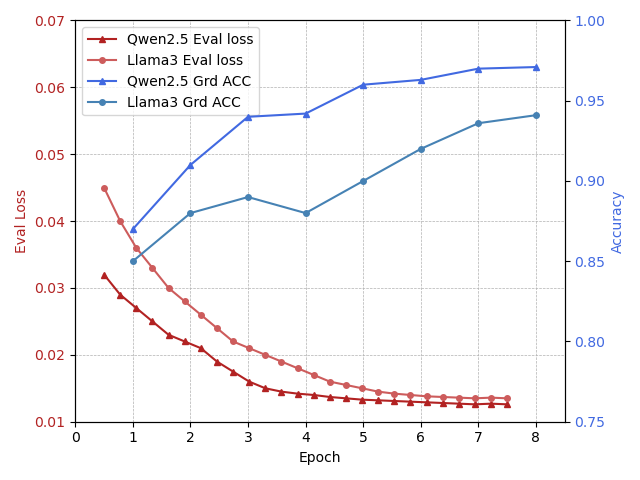}
    \caption{Learning curves of semantics prediction.}
    \label{fig_curve}
\end{figure}

\subsection{Metrics}
To evaluate the performance of the models, we consider two perspectives: the extension view and the argument view.

\paragraph{Extension View Accuracy}
From the extension view, we define the accuracy of extension prediction, denoted as ACC$_{\sigma}$. 
This metric represents the proportion of cases where the predicted $\mathtt{IN}(\mathcal{L})$ exactly matches the ground truth extension for a given semantics $\sigma$. 
We also use the Pass@k metric to evaluate model robustness in predicting extensions.

\paragraph{Argument View Accuracy}

From the argument view, we assess the acceptability of arguments by calculating their credulous and skeptical acceptability based on the predicted labellings. For argument acceptability classification, we use the Matthews Correlation Coefficient (MCC), which is particularly suited for imbalanced datasets. The metrics are denoted as MCC$_{c}$ for credulous acceptability and MCC$_{s}$ for skeptical acceptability, respectively.

To evaluate the effectiveness of conflict resolution, we employ the following two metrics:

\paragraph{Conflict-Free Proportion} 
This metric measures the average proportion of generated extensions that satisfy the \textit{conflict-freeness} property across all argumentation frameworks in the dataset. The conflict-freeness property ensures that no two arguments in a single extension attack each other. Formally, for a dataset with $N$ argumentation frameworks, the Conflict-Free Proportion (CFP) is defined as:
\[
\text{CFP} = \frac{1}{N} \sum_\mathcal{E} \frac{|\{E \in \mathcal{E} \mid E \text{ is conflict-free}\}|}{|\mathcal{E}|}
\]
where $\mathcal{E}$ represents the set of generated extensions. 

\paragraph{At Least One Successful Extension} 
The metric evaluates whether at least one of the generated extensions is valid. For a given semantics $\sigma$ and a set of generated extensions $\mathcal{E}$, the At Least One Successful Extension (ALSE) for a single argumentation framework $\mathcal{F}$ is defined as:
\[
\text{ALSE}_\mathcal{F}(\mathcal{E}) = 
\begin{cases} 
1, & \text{if } \exists E \in \mathcal{E}_\mathcal{F} \text{ such that } E \in \sigma(\mathcal{F}), \\
0, & \text{otherwise}.
\end{cases}
\]
The overall ALSE score for a dataset is then computed as the proportion of cases where at least one valid extension is generated:
\[
\text{ALSE} = \frac{1}{N} \sum_\mathcal{F} \text{ALSE}_\mathcal{F}(\mathcal{E}_\mathcal{F})
\]
where $\mathcal{E}_\mathcal{F}$ represents the set of generated extensions for the framework $\mathcal{F}$.

\subsection{Main Results}

In Table~\ref{tab_main}, we present the performance metrics of various language models, categorized by different semantics. 
For grounded semantics, the credulous and skeptical acceptance of arguments are identical, so they are combined into a single performance metric. 
Similarly, the credulous acceptances under both complete and preferred semantics are identical, resulting in the same MCC scores. 
It should be noted that complete semantics (and thus preferred/stable semantics) are computed with the assumption that grounded semantics is already known.

The upper section of Table~\ref{tab_main} presents the zero-shot CoT approaches, encompassing the advanced GPT, Llama, and Qwen series models. 
These results highlight the models' initial performance without prior specific training. 
This baseline assessment is mainly for understanding their inherent capabilities in handling different semantic frameworks.
In contrast, the lower section of the table showcases the outcomes after applying our training methodology. 
The final two rows demonstrate significant improvements in performance, underscoring the effectiveness of our approach. 
This enhancement illustrates the model's ability to adapt and optimize through training.

It is worth noting that the accuracy of our models in \prf\ prediction is comparable to, but does not exceed, that of GNN-based approaches~\cite{ijcai2020p231}, which benefit from architectures well-suited to argumentation tasks.
However, a key advantage of our method lies in its ability to directly compute extensions and provide corresponding explanations, offering a comprehensive understanding of the argumentation process. 
This capability underscores the practical applicability of our model in diverse semantic contexts.

The performance variance in the training is plotted in Figure~\ref{fig_curve},
illustrating the cross-entropy loss and accuracy as a function of training epochs. 
The red curves represent the evaluation loss, while the blue curves denote accuracy. 
These plots provide insight into the training dynamics and convergence behavior of the models over time.

\subsection{Effectiveness on Conflict Resolving}

\begin{table}[t]
\begin{center}
\scalebox{0.9}[0.9]{
\begin{tabular}{lcccccccc}
\toprule
\multirow{2}{*}{Models} & \multicolumn{2}{c}{\com} && \multicolumn{2}{c}{\prf} \\
\cline{2-3} \cline{5-6} 
 &               CFP    & ALSE  && CFP  & ALSE
\cr
\hline
Qwen2.5-7B        & 0.71  & 0.62  && 0.78 & 0.63  \\
Llama3-8B       & 0.51  & 0.42  && 0.45 & 0.48  \\
\hline
\specialrule{0em}{1pt}{1pt}
Qwen2.5-SFT     & 1.00 & 0.99 && 1.00  & 0.95  \\
Llama3-SFT   & 1.00 & 0.99 && 1.00  & 0.86  \\
\bottomrule
\end{tabular}}
\end{center}
\caption{The conflict resolving ability comparison of LLMs.}
\label{tab_conflict}
\end{table}

The evaluation results in Table~\ref{tab_conflict} notably demonstrate the differences in conflict resolution capabilities before and after training.
Before training, Qwen2.5B achieves CFP scores of 0.71 and 0.78, and ALSE scores of 0.62 and 0.63 under the complete and preferred semantics, respectively.
In comparison, Llama3 shows lower performance, with CFP scores of 0.51 and 0.45, and ALSE scores of 0.42 and 0.48 in the same scenarios.
After training, both models achieve near-perfect results: Qwen2.5-SFT attains CFP and ALSE scores of 1.00 and 0.99 under \com, and 1.00 and 0.95 under \prf; while Llama3-SFT reaches 1.00 and 0.99 under \com, and 1.00 and 0.86 under \prf.
These results clearly demonstrate that training significantly enhances the models' abilities to generate logically consistent and semantically valid solutions in argumentation frameworks.

\subsection{Robustness}
\begin{figure}
    \centering
    \includegraphics[width=1\linewidth]{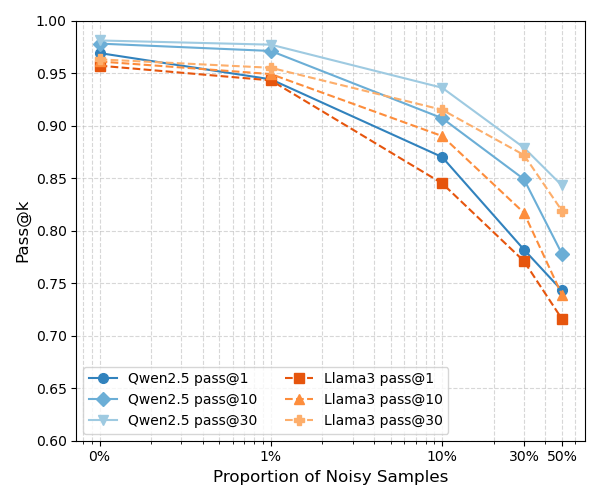}
    \caption{Robustness: Pass@k vs. Noisy Data Ratio}
    \label{fig_robust}
\end{figure}

We generate our dataset through symbolic computation to ensure clean data.
To evaluate the robustness of our trained models against noise, we manually create corrupted datasets by generating erroneous samples for grounded semantics.
Specifically, to generate an error case, we randomly select an intermediate set in a clean derivation and replace half of its elements with random values.
For example, we may modify the label set \texttt{IN} from $\{1,3,6,7\}$ to $\{2,3,5,7\}$, and then recompute \texttt{OUT} and \texttt{UNDEC} accordingly.
Such a step error can propagate to the final answer, thereby introducing noise into the reasoning process.

To assess the stability of model outputs, we train the models on the corrupted datasets and report Pass@k performance in Figure~\ref{fig_robust} as the proportion of noisy samples increases.
As shown in the results, both models exhibit a gradual decrease in Pass@k as the proportion of noisy samples increases, and Qwen2.5 consistently outperforms Llama3 under all noise levels.

\subsection{Direct vs. Indirect Complete Semantics Computation}
\begin{figure}[t]
    \centering
    \includegraphics[width=0.8\linewidth]{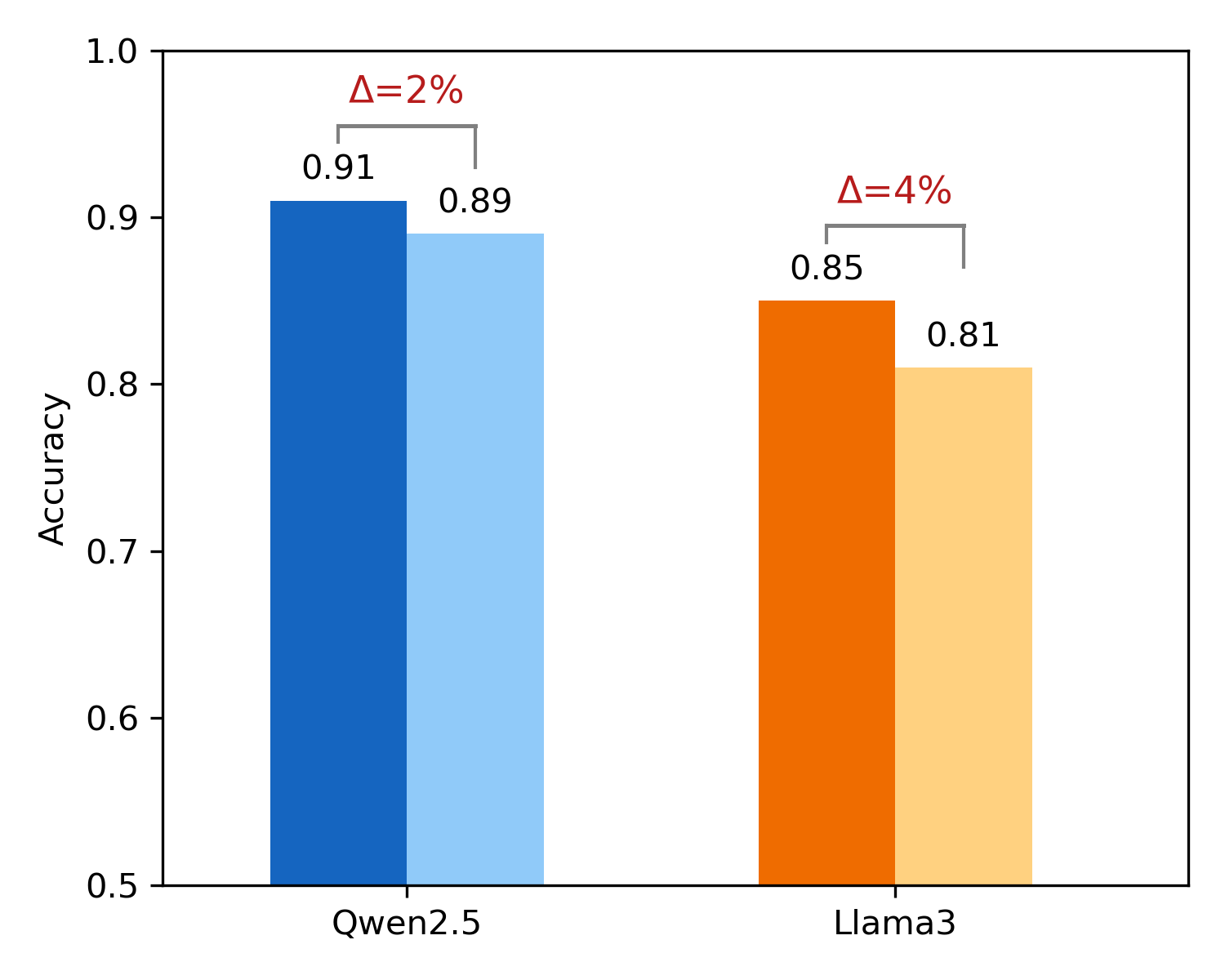}
    \caption{Accuracy of two-step vs. end-to-end prediction of complete extensions.}
    \label{fig_e2e}
\end{figure}

In the task of predicting the complete extensions, we divide the computation into two steps: first computing the grounded extension, and then computing the complete extension. 
In the second step, we provide the grounded extension as a known condition in the prompt. 
If we adopt an end-to-end approach to predict the complete extensions, errors can accumulate across steps. 
In real-world applications, however, such end-to-end prediction is often more meaningful.
Figure~\ref{fig_e2e} shows the results for both methods. For each model, the left bar represents the two-step approach, while the right bar represents the end-to-end approach.
The annotated gap $\Delta$ indicates the accuracy drop caused by error accumulation in the end-to-end setting.

\subsection{Effect of Graph Input}

AAF structures can be visualized as directed graphs for model input (see Figure~\ref{fig_aaf1}).
To investigate the effect of visualized graph inputs, we conduct experiments using Qwen2.5-VL-7B~\cite{qwen25vl}, a multi-modal model comparable in scale to Qwen2.5-7B.
For comparison, we evaluate both unimodal (text-only) and multimodal (visual+text) settings.
In the Visual+Text setting, the problem description is converted into an image, while other components remain unchanged.
Table~\ref{tab_vl} summarizes the performance under different input formats.
For the Qwen2.5-VL-SFT model, using visualized graph inputs together with text achieves slightly higher accuracy on both grounded extension and complete extension prediction compared to the text-only setting.
This suggests that incorporating visual representations of AAF structures can provide additional benefits in understanding and reasoning, although the improvement is marginal in this case.

\subsection{Errors in Computing Grounded Semantics}
In this part, we summarize the errors and failures encountered by the LLMs when computing the grounded labelling.

\begin{table}[t]
\centering
\label{tab_vl}
\begin{tabular}{lccccc}
\toprule
Model           & Input Format  & ACC$_\grd$ & ACC$_\com$ \\
\midrule
Qwen2.5-7B      & Text          & 0.31       & 0.43    \\
Qwen2.5-SFT     & Text          & 0.97       & 0.85    \\
Qwen2.5-VL-SFT  & Text          & 0.98       & 0.87     \\
Qwen2.5-VL-SFT  & Visual+Text   & 0.98       & 0.88    \\
\bottomrule
\end{tabular}
\caption{Performance comparison under different input formats.}
\end{table}

\begin{figure*}[t]
    \centering
    \includegraphics[width=1\linewidth]{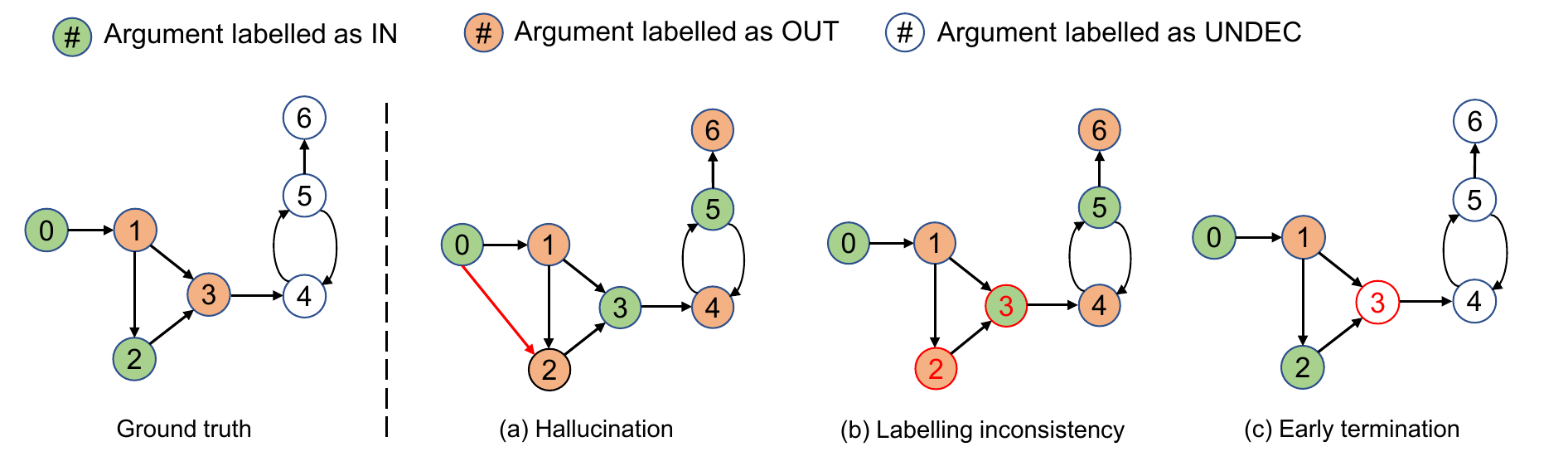}
    \caption{Several errors made by LLMs when computing the grounded labelling. Erroneous parts are highlighted in red.}
    \label{fig_err1}
\end{figure*}

\begin{figure}
    \centering
    \includegraphics[width=1\linewidth]{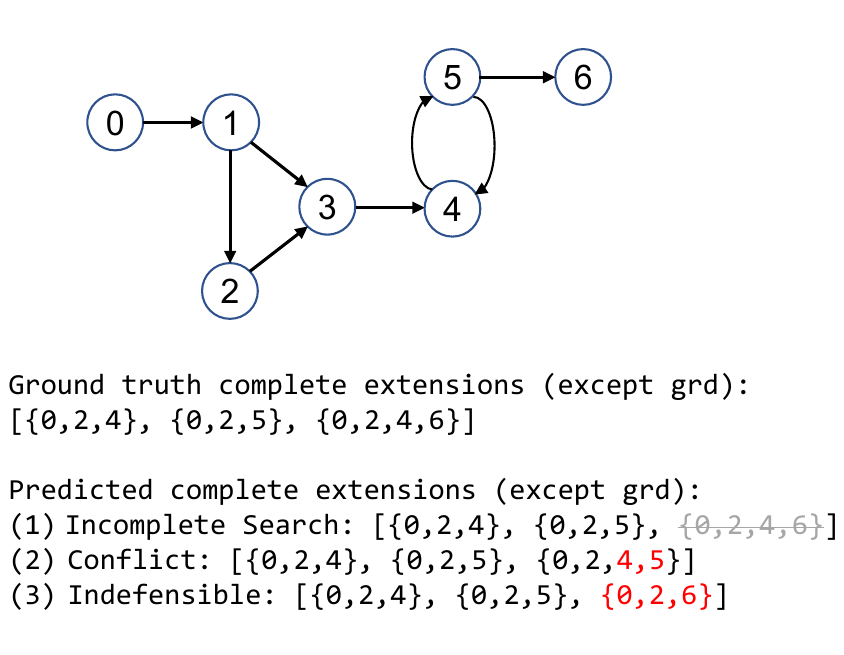}
    \caption{Errors encountered by LLMs when computing the complete extensions. Erroneous parts are highlighted in red.}
    \label{fig_err2}
\end{figure}

\subsubsection{Hallucination}
Hallucination typically refers to the generation of meaningless or unfaithful information that does not adhere to the provided source content.
In abstract argumentation tasks, hallucinations in LLMs typically lead to the generation of information or text that is inconsistent with the input, thereby resulting in errors in the model's resolution process.
For instance, in Figure~\ref{fig_err1}~(a), the LLM incorrectly references the attack ``2 is attacked by 0 which is labelled IN'', 
while this attack actually does not exist.

\subsubsection{Labelling Inconsistency}
Although LLMs correctly generate arguments' labels, in subsequent steps, the model may erroneously reinterpret these labels, leading to incorrect solutions. 
As shown in Figure~\ref{fig_err1}~(b), the argument 2 is mistakenly presumed to have been assigned $\mathtt{OUT}$, when it should actually be assigned $\mathtt{IN}$.

\subsubsection{Early Termination}
During the iterative computation phase, the model prematurely ends the computation process without assigning all arguments that could potentially be made $\mathtt{IN}$ or $\mathtt{OUT}$. 
For example, in Figure~\ref{fig_err1}~(c), the computation ceases after the IN step, specifically after assigning $\mathtt{IN}$ to 2, and fails to initiate the subsequent OUT step. 
This abrupt termination prevents the assignment of $\mathtt{OUT}$ to 3, of with the label remains $\mathtt{UNDEC}$.

\subsubsection{Wrong Label Assignment}
The grounded extension is computed by alternately assigning $\mathtt{IN}$ and $\mathtt{OUT}$ to arguments. However, errors can occur if the sequence is not followed. In Figure~\ref{fig_err1}, an error occurs when the LLM mistakenly assigns $\mathtt{IN}$ to 4, based on the incorrect assumption that 3 and 5, which attack 4, are assigned $\mathtt{OUT}$. This error results from the incorrect assignment of $\mathtt{OUT}$ to 5, which is attacked by 4, incorrectly labeled as $\mathtt{IN}$.

\subsection{Errors in Computing Complete Semantics} 
In the computation of complete extensions, LLMs exhibit several critical inconsistencies:
\subsubsection{Incomplete Search} 
LLMs occasionally fail to identify all potential extensions, resulting in an incomplete coverage of the complete extensions. For example, in Figure~\ref{fig_err2}~(1), the predicted complete extensions lack \{0, 2, 4, 6\}.
\subsubsection{Conflict} 
During the inclusion of arguments labelled as $\mathtt{IN}$, LLMs may incorrectly assess these arguments as being conflict-free. For example, in Figure~\ref{fig_err2}~(2), 4 and 5 are added to the $\mathtt{IN}$ set even though they mutually attack each other.
\subsubsection{Indefensibility} 
The complete extension generated by LLMs may not always defend all its arguments. For example, in Figure~\ref{fig_err2}~(3), \{0,2,6\} does not defend itself from 5.

\subsection{Error Analysis of the GNN Baseline} 

\begin{figure}
    \centering
    \includegraphics[width=0.75\linewidth]{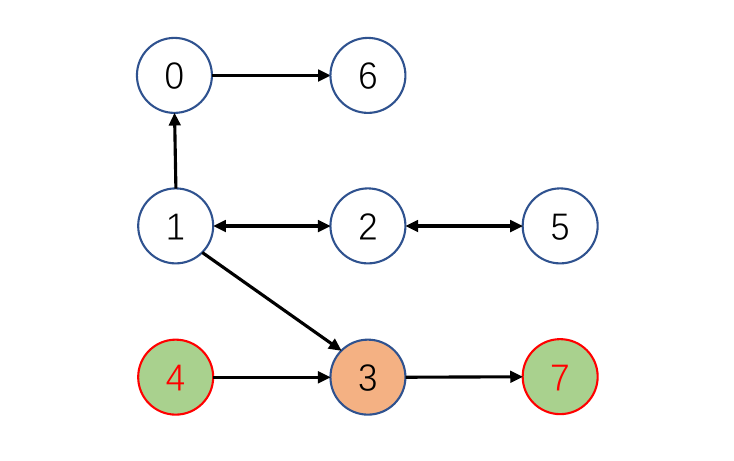}
    \caption{A failure case of AGNN: The model incorrectly predicts the grounded extension as the empty set, while the correct answer is \{4, 7\}.}
    \label{fig_err3}
\end{figure}

To compare the error patterns of GNNs and our LLM-based approach, we analyze the failure cases in AGNN~\cite{ijcai2020p231}.
Our in-depth examination reveals a prevalent failure mode that does not exist in LLMs: in the vast majority of incorrect cases, the model fails to infer a valid grounded extension and instead incorrectly predicts the empty set, even when a correct extension could be easily found.
An example of such a failure is shown in Figure~\ref{fig_err3}.
Because GNNs learn from data, the reasoning process behind their predictions lacks transparency, making it difficult to interpret intuitively and thereby reducing our confidence in the model's predictions in these specific failure cases.

\section{Empirical Study on Process Explanations}

In this section, we examine the effect of explanations on model performance and generalization capability. 
We find that within the existing data distribution ($n \leq 25$), explanations have only a marginal impact on performance.
While when applied to AAFs with more arguments, which we referred to as the generalization stage, models trained with explanations demonstrate stronger generalization ability.
Additionally, we observe that human-annotated explanations further enhance the diversity of model-generated explanations, leading to better explanatory quality, which is important for human-AI interaction in real-world applications.

\subsection{Role of Explanations in Generalization}

To demonstrate the efficacy of explanations in data generation, as a comparison model, we additionally trained models, namely Llama3-noexp and Qwen2.5-noexp, on the datasets without explanations, where the output text contained only the answer.
To assess generalization capability across varying numbers of arguments, we generate additional test datasets spanning from 26 to 35 in number, each comprising 100 samples.

\begin{figure}[t]
    \centering
    \includegraphics[width=1\linewidth]{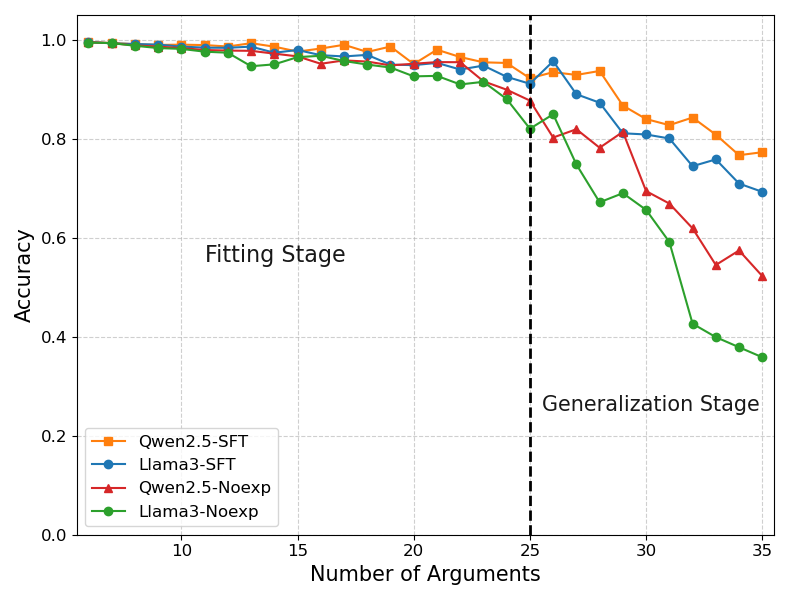}
    \caption{Accuracy variance of grounded extension prediction across the number of arguments.}
\label{fig_scaling}
\end{figure}

For the grounded extension prediction,
Figure~\ref{fig_scaling} illustrates the accuracy variance of these models across subsets differentiated by the number of arguments. 
We delineate the plot into two stages: \textit{the fitting stage and the generalization stage}, demarcated by a vertical dashed line. Performance in the fitting stage corresponds to model fitting ability, while the generalization stage elucidates their capacity to extrapolate to novel scenarios.

\subsubsection{Fitting Stage}
During the fitting stage, models exhibit high accuracy and relatively stable performance, demonstrating effective learning of the training dataset. 
This is attributed to the consistent data distribution, allowing models to perform robustly. 
Models trained with explanations show a slight advantage, suggesting that explanations contribute to more robust internal representations even at this stage.

\subsubsection{Generalization Stage}
In the generalization stage, significant differences emerge between models trained with and without explanations. 
Models with explanations maintain relatively high accuracy, though performance gradually declines as complexity increases.
Conversely, models without explanations experience a sharp drop in accuracy, sometimes falling to 30\%, indicating poor generalization. 
This disparity underscores the importance of explanations in facilitating the acquisition of underlying logical structures, enhancing model robustness and adaptability to novel, complex scenarios. 
These findings emphasize the need for incorporating explanatory components in training to improve model resilience and generalization.

\subsection{Ablation Study}

\begin{table}[t]
\centering
\label{tab_abl}
\begin{tabular}{lccccc}
\toprule
Model           & Exp. Component  & ACC$_\grd$ & ACC$_\com$ \\
\midrule
Qwen2.5-SFT     & None          & 0.68       & 0.56    \\
Llama3-SFT      & None          & 0.58       & 0.42     \\
Qwen2.5-SFT     & \textsf{Q}    & 0.72       & 0.59     \\
Llama3-SFT      & \textsf{Q}    & 0.66       & 0.53    \\
Qwen2.5-SFT     & \textsf{P}$\rightarrow$\textsf{Q}  & 0.85   & 0.78    \\
Llama3-SFT      & \textsf{P}$\rightarrow$\textsf{Q}  & 0.80   & 0.71     \\
\bottomrule
\end{tabular}
\caption{Ablation study on explanation components.}
\end{table}

In the computation of grounded extensions, we describe each step in logical form as explanations.
The explainability of data samples primarily arises from process-level reasoning at each step, where the explanation is presented as ``\textsf{P}, so \textsf{Q}.''
Our previous experiments have shown that training with explanations can endow models with greater generalization ability. 
Here, we aim to investigate whether this improved generalization is specifically guided by the underlying logical relationships.
To perform this ablation study, we mask \textsf{P} to hide the explanation effect, or mask both \textsf{P} and \textsf{Q} to further obscure the reasoning process.
Table~\ref{tab_abl} reports the averaged accuracy of extension prediction in the generalization stage (i.e., for the number of arguments greater than 25).
Notably, including the full logical explanation (\textsf{P}$\rightarrow$\textsf{Q}) leads to significantly better generalization performance compared to masking parts of the explanation.
In contrast, providing only the result (\textsf{Q}) at each step yields lower accuracy.
These results suggest that explicit logical reasoning steps in explanations play a crucial role in improving model generalization.

\subsection{Effect of Human Feedback}

\begin{figure}
    \centering
    \includegraphics[width=0.8\linewidth]{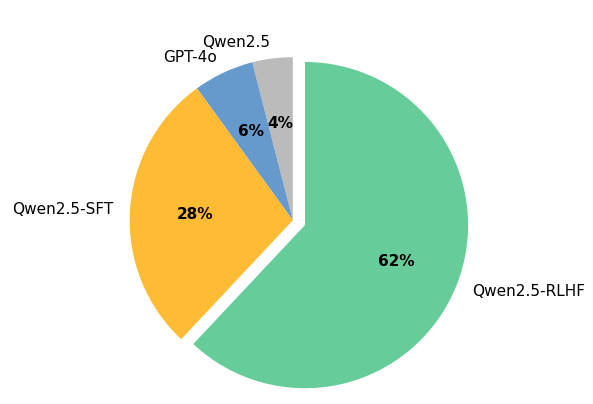}
    \caption{Participant preferences for process explanations.}
    \label{fig_rlhf}
\end{figure}

In the subsequent evaluation phase, we invited non-expert participants to perform pairwise preference tests, comparing explanations generated by four different models: Qwen2.5-7B, GPT-4o, Qwen2.5-SFT, and Qwen2.5-RLHF. 
For each of 50 randomly selected questions from the test set, 10 participants were asked to choose the explanation they found most understandable or helpful, providing a human-centered evaluation of explainability.
Note that model outputs may contain errors, and participants are instructed to take these into account as part of their evaluation.
The results are summarized in Figure~\ref{fig_rlhf}.
Our findings suggest that integrating human feedback via RLHF can substantially improve the explainability of model-generated decisions, making them align more closely with human reasoning and teaching practices.

\section{Limitations and Future Work}

Our training approach primarily focuses on abstract argumentation, enabling LLMs to resolve conflicts in a more logical manner. However, we do not explore the process of grounding real-world scenarios to abstract models. 
Mapping complex real-world situations to arguments and attack/support relations is another important issue, where fairness and representational bias may arise~\cite{chen2024credible,cramer2019empirical}. 
Encoding inductive biases into formal logical structures can directly impact the applicability of abstract argumentation in LLMs.
Additionally, argumentation graphs derived from complex real-world situations may exhibit greater structural complexity, and our current data distribution may not comprehensively cover such edge cases.

There are two promising directions to address these key challenges. The first is to train models that can map conflict-rich real-world scenarios to formal representations, thereby improving the connection between practical applications and formal methods. The second is to leverage richer argumentation models to enhance the model's ability to handle uncertain reasoning, including probabilistic settings~\cite{hunter2021probabilistic} and quantitative argumentation frameworks~\cite{potyka2021interpreting}.

\section{Conclusion}
In this paper, we examine the computation of abstract argumentation frameworks utilizing large language models. 
Our approach integrates argumentation computation with the capabilities of LLMs, providing enhanced explainability over traditional graph neural network methods. 
Experimental results demonstrate that LLMs can effectively learn and execute extension computation tasks. 
Key findings highlight the significant role of explanations in enhancing model performance and generalization. 
The self-explanatory capacities of LLMs address transparency challenges typically associated with neural networks. 
This research advances the understanding of LLMs in argumentation computation and paves the way for further investigation into defeasible reasoning and complex decision-making processes.
Our future works include exploring how LLMs can adapt to dynamic and incomplete argumentation frameworks, where arguments and relationships evolve or are missing, would enhance their applicability to real-world scenarios like legal reasoning and ethical decision-making.

\section*{Acknowledgments} 
This research was partially supported by the National Natural Science Foundation of China (62576309).

\bibliographystyle{elsarticle-num} 
\bibliography{cas-refs}

\clearpage
\appendix
\twocolumn[
\begin{center}
    \Huge Appendix
\end{center}
\vspace{1cm}
]

\section{Chain-of-Thought Baseline}

\begin{figure*}[t]
    \centering
    \includegraphics[width=0.95\linewidth]{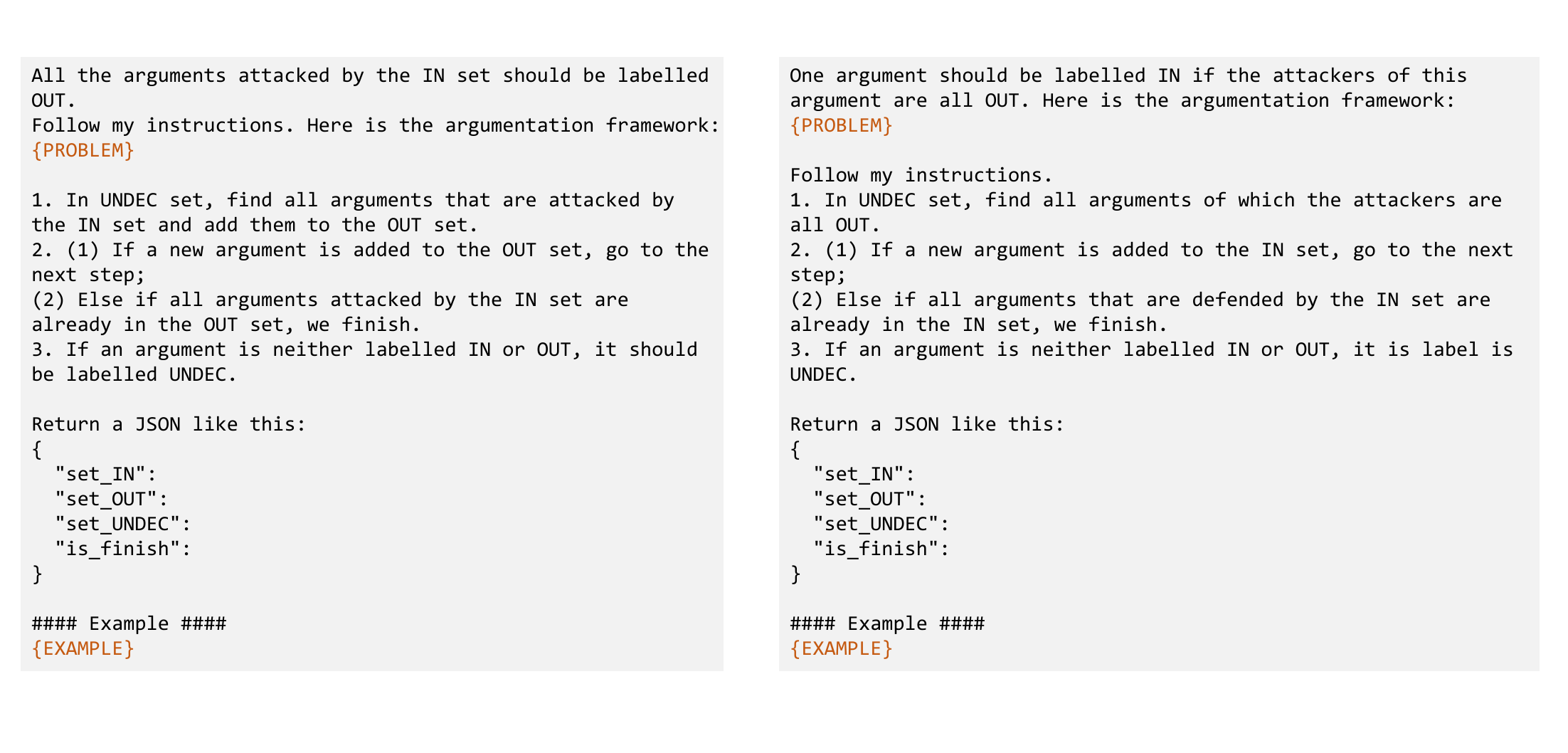}
    \caption{Prompt of the recursive part of the algorithm.}
    \label{fig_prompt1}
\end{figure*}

\begin{figure*}[t]
    \centering
    \includegraphics[width=0.95\linewidth]{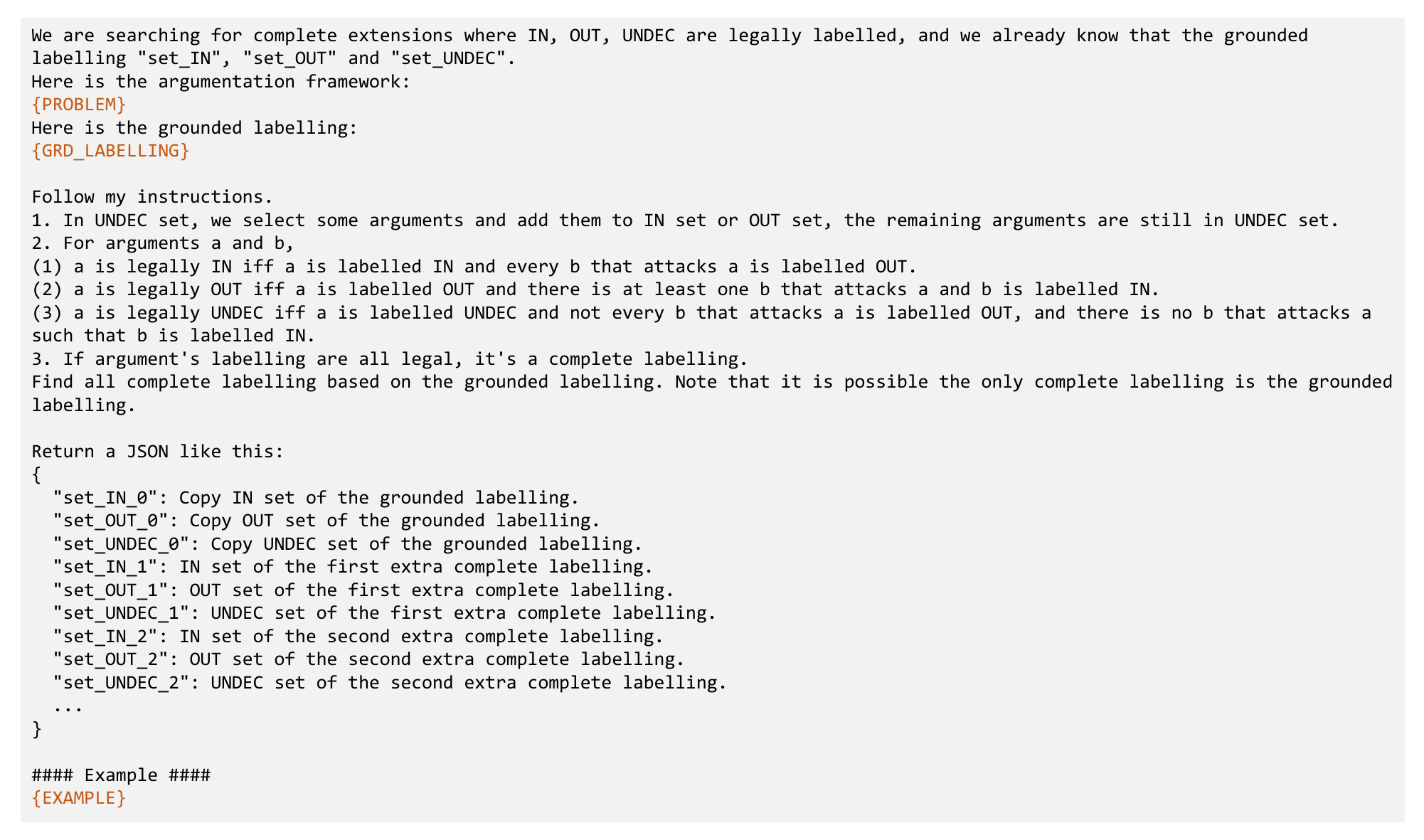}
    \caption{Prompt of computing complete labellings.}
    \label{fig_prompt2}
\end{figure*}

Prompting generally results in models making predictions with lower accuracy compared to models that have been fine-tuned with substantial training data~\cite{zhang2023large}. 
To validate this observation and establish a baseline method, we design a Chain-of-Thought (CoT) pipeline~\cite{NEURIPS2022_9d560961} that computes extensions based on a labelling algorithm. 
The argumentation computation is divided into multiple steps, framed as a reasoning task.

\subsubsection{Grounded Semantics}
Initially, we employ LLMs to extract a list of arguments and identify attack relation to address the problem. 
We then task the LLMs with finding unattacked arguments to initialize the set $\mathtt{IN}(\mathcal{L})$. 
By following each step in the recursive labelling algorithm, the LLMs ultimately determine the grounded labeling when no further arguments can be added to $\mathtt{IN}(\mathcal{L})$ or $\mathtt{OUT}(\mathcal{L})$. 
The algorithm running prompt template is illustrated in Figure~\ref{fig_prompt1}, where the special placeholders ``\{PROGRAM\}'' and ``\{EXAMPLE\}'' will be replaced by specific data sample.

\subsubsection{Complete Semantics}
The subsequent task involves identifying complete labellings based on the grounded labelling.  
This process begins by selecting elements from $\mathtt{UNDEC}(\mathcal{L})$ and reassigning them to $\mathtt{IN}$ or $\mathtt{OUT}$. 
Similar to the method used in model training, LLMs are directed to propose solutions and subsequently verify the legality of these choices, as shown in Figure~\ref{fig_prompt2}.
The special placeholder ``\{GRD\_LABELLING\}'' will be replaced by ground truth grounded labelling.

\begin{figure*}[t]
    \centering
    \includegraphics[width=0.8\linewidth]{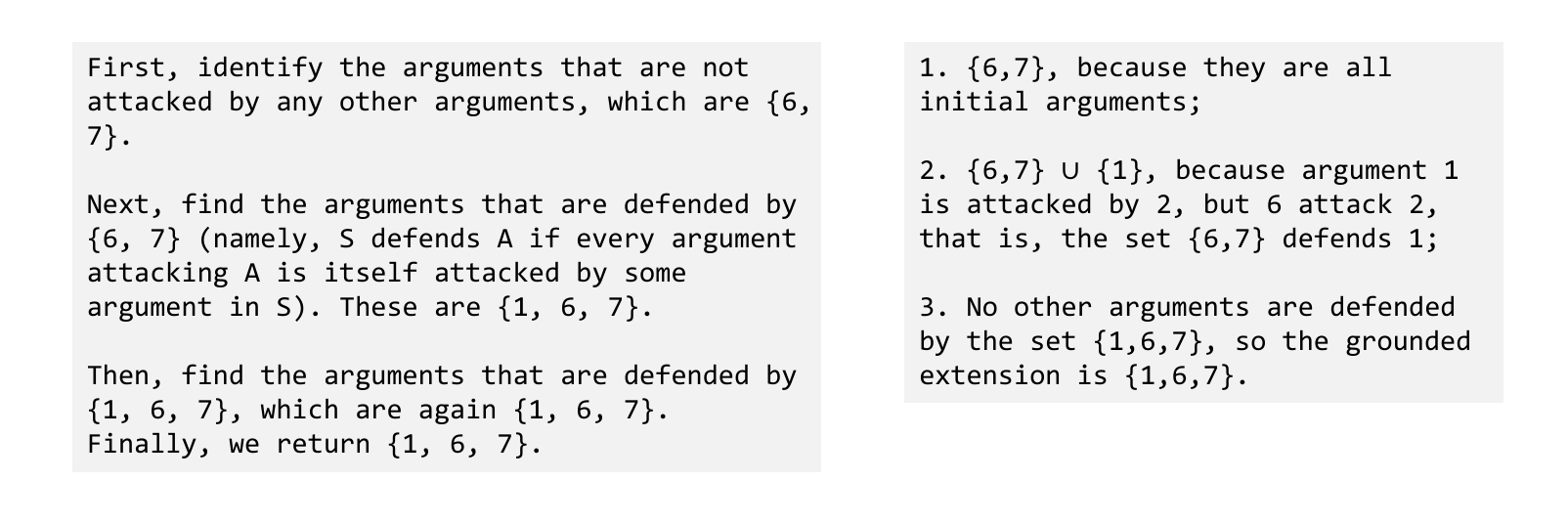} 
    \caption{Examples of two human annotations of explanation.}
    \label{fig_human_anno}
\end{figure*}

\section{Integrating Human Feedback}

For human-annotated explanations, due to the limited availability of expert annotators, we focus on explanations for the computation of the grounded extension and annotate a subset of high-quality data. 
For each number of arguments, we annotate 50 distinct problems, resulting in a total of 1,000 training examples. 
For RLHF training, we further fine-tune the SFT-trained model using the GRPO algorithm~\cite{shao2024deepseekmath} within the TRL~\cite{vonwerra2022trl} framework, leveraging the collected human-annotated explanations as feedback. 
Figure~\ref{fig_human_anno} presents two examples of human annotations.
To ensure the model outputs only the extension (rather than the label as in SFT), we manually add a prompt at the end requesting a JSON object containing the extension set.
The reward function is defined as follows: a score of 1 is assigned if the extension is completely correct, and 0 otherwise.

\end{document}